\documentclass{bmvc2k}

\title{Weakly supervised cross-domain alignment with optimal transport}

\addauthor{Siyang Yuan}{siyang.yuan@duke.edu}{1}
\addauthor{Ke Bai}{ke.bai@duke.edu}{1}
\addauthor{Liqun Chen}{liqun.chen@duke.edu}{1}
\addauthor{Yizhe Zhang}{yizzhang@microsoft.com}{2}
\addauthor{Chenyang Tao}{chenyang.tao@duke.edu}{1}
\addauthor{Chunyuan Li}{lichunyuan24@gmail.com}{2}
\addauthor{Guoyin Wang}{guoyinwang.duke@gmail.com}{3}
\addauthor{Ricardo Henao}{rocardo.henao@duke.edu}{1}
\addauthor{Lawrence Carin}{lcarin@duke.edu}{1}

\addinstitution{
 Duke University\\
 Durham, North Carolina, USA
}
\addinstitution{
 Microsoft Research\\
 Redmond, Washington, USA
}
\addinstitution{
 Amazon, Alexa AI\\
 Seattle, Washington, USA
}

\runninghead{Yuan et al.}{weakly supervised cross-domain alignment with OT}

\usepackage{epsfig}
\usepackage{graphicx}

\usepackage{booktabs}
\usepackage{wrapfig}
\usepackage{algorithm}
\usepackage{algorithmic}
\usepackage{amsmath}

\newcommand{\BR}{\mathbb{R}}

\newcommand{\CX}{\mathcal{X}}

\usepackage{amsmath,amsfonts,bm}

\newcommand{\ie}[0]{\emph{i.e., }}

\def\eqref#1{equation~\ref{#1}}

\def\1{\bm{1}}

\def\vv{{\bm{v}}}

\DeclareMathAlphabet{\mathsfit}{\encodingdefault}{\sfdefault}{m}{sl}
\SetMathAlphabet{\mathsfit}{bold}{\encodingdefault}{\sfdefault}{bx}{n}

\def\sR{{\mathbb{R}}}

\newcommand{\R}{\mathbb{R}}

\DeclareMathOperator*{\argmax}{arg\,max}

\newcommand{\Cmat}{{\bf C}}

\newcommand{\Emat}{{\bf E}}
\newcommand{\Fmat}{{\bf F}}

\newcommand{\Imat}{{\bf I}}

\newcommand{\Smat}{{\bf S}}
\newcommand{\Tmat}{{\bf T}}

\newcommand{\Vmat}{{\bf V}}
\newcommand{\Wmat}{{\bf W}}

\newcommand{\bv}{{\boldsymbol b}}

\newcommand{\ev}{{\boldsymbol e}}
\newcommand{\fv}{{\boldsymbol f}}

\newcommand{\hv}{{\boldsymbol h}}

\newcommand{\wv}{{\boldsymbol w}}

\newcommand{\xv}{{\boldsymbol x}}
\newcommand{\yv}{{\boldsymbol y}}

\newcommand{\muv}{{\boldsymbol \mu}}
\newcommand{\nuv}{{\boldsymbol \nu}}

\newcommand{\Lcal}{\mathcal{L}}

\newcommand{\Dcal}{\mathcal{D}}

\begin{document}
\maketitle

\begin{abstract}
Cross-domain alignment between image objects and text sequences is key to many visual-language tasks, and it poses a fundamental challenge to both computer vision and natural language processing.
This paper investigates a novel approach for the identification and optimization of  fine-grained semantic similarities between image and text entities,
under a weakly-supervised setup, improving performance over state-of-the-art solutions.
Our method builds upon recent advances in optimal transport (OT) to resolve the cross-domain matching problem in a principled manner. Formulated as a drop-in regularizer, the proposed OT solution can be efficiently computed and used in combination with other existing approaches. 
We present empirical evidence to demonstrate the effectiveness of our approach, showing how it enables simpler model architectures to outperform or be comparable with more sophisticated designs on a range of vision-language tasks.

\end{abstract}
\vspace{-4mm}
\section{Introduction}
\vspace{-1mm}
The intersection between computer vision (CV) and natural language processing (NLP) has inspired some of the most active research topics in artificial intelligence.
Prominent examples of such work includes image-text retrieval~\cite{karpathy2015deep, SCAN}, image captioning~\cite{xu2015show, karpathy2015deep, fang2015captions, vinyals2015show, johnson2016densecap}, text-to-image generation ~\cite{reed2016generative, qiao2019mirrorgan}, phrase localization ~\cite{Align2ground, plummer2015flickr30k} and visual question answering (VQA)~\cite{antol2015vqa, malinowski2014multi}.
Core to these applications is the challenge of {\it cross-domain alignment} (CDA), consisting of accurately associating related entities across different domains in a cost-effective fashion.

Contextualized in image-text applications, the goal of CDA is two-fold: 
$i$) identify entities in images ({\it e.g.}, regions or objects) and text sequences ({\it e.g.}, words or phrases); and then $ii$) quantify the relatedness between identified cross-domain entity pairs.
CDA is particularly challenging because it constitutes a {\it weakly supervised learning task}.
More specifically, neither the entities nor their correspondence ({\it i.e.}, the match between cross-domain entities) is labeled~\cite{harwath2018jointly}.
This means that CDA must learn to identify entities and quantify their correspondence only from the image-text pairs during training.

Given the practical significance of CDA, considerable effort has been devoted to address this challenge in a scalable and flexible fashion. Existing solutions often explore heuristics to design losses that encode cross-domain correspondence. 
Pioneering investigations, such as ~\cite{kiros2014unifying}, considered entity matching via a hinge-based ranking loss applied to shared latent features of image and text, extracted respectively with a convolutional neural network (CNN)~\cite{lecun1999object} and long short-term memory (LSTM)~\cite{hochreiter1997long} feature encoders.
Explicitly modeling the between-entity relations also yields significant improvements \cite{karpathy2015deep}. Performance gains can also be expected via exploiting the hardest negatives in a triplet ranking loss specification~\cite{VSE}.
More recently, synergies between CDA and attention mechanisms~\cite{nam2017dual} have been explored, further advancing the state of the art with more sophisticated model designs \cite{SCAN}.

Despite recent progress, it remains an open question concerning which other (mathematical) principles can be leveraged for scalable automated discovery of cross-domain relations.
This study develops a novel solution based on recent developments in {\it optimal transport} (OT) based learning~\cite{cuturi2017computational}.
Briefly, OT-based learning is a generic framework that seeks to tackle specific problems by recasting them as distribution matching problems, which can then be accurately and efficiently solved by optimizing the transport distance between the distributions.
Its recent success in addressing fundamental challenges in artificial intelligence has sparked a surge of interest in extending its reach to other applications~\cite{arjovsky2017wasserstein,liu2018scene,chen2019improving}.

Our work is motivated by the insight that cross-domain alignment can be reformulated as a bipartite matching problem \cite{kuhn1955hungarian}, which can be optimized w.r.t. a proper matching score. 
We show that a solution to the challenge of automated cross-domain alignment can be approached by using the optimal transport distance as the matching score. 
Notably, our construction is orthogonal to the development of cross-domain attention scores~\cite{nam2017dual,yu2017multi,SCAN}, which are essentially advanced feature extractors \cite{bahdanau2014neural} and cannot be used as an optimization criteria for the purpose of CDA {\it per se}, necessitating a pre-specified objective function in training for feature alignment.
For example, in image captioning, maximum likelihood estimation (MLE) is applied to match generated text sequences to the reference (ground truth), and in image-text retrieval the models are typically optimized w.r.t. their ranking~\cite{SCAN}.
In this sense, the learning of attention scores is guided by MLE or ranking loss, while our OT objective can be directly optimized during training to learn optimal matching strategies.

\begin{figure}[t!]
 \centering
  \begin{subfigure}
    \centering
    \includegraphics[height=30mm]{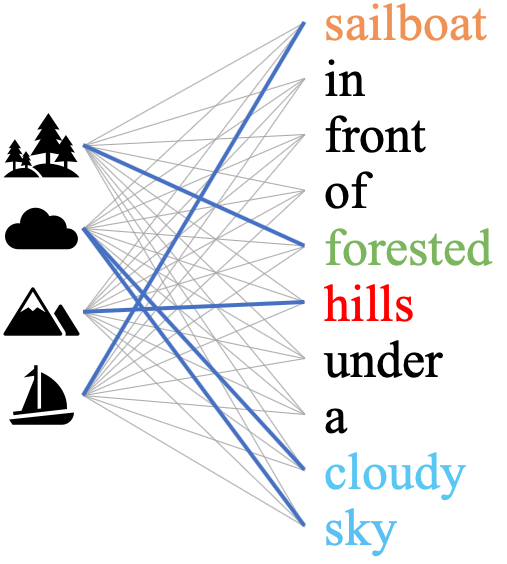}
    \end{subfigure}
    \hspace{3mm}
 \begin{subfigure}
    \centering
    \includegraphics[height=30mm]{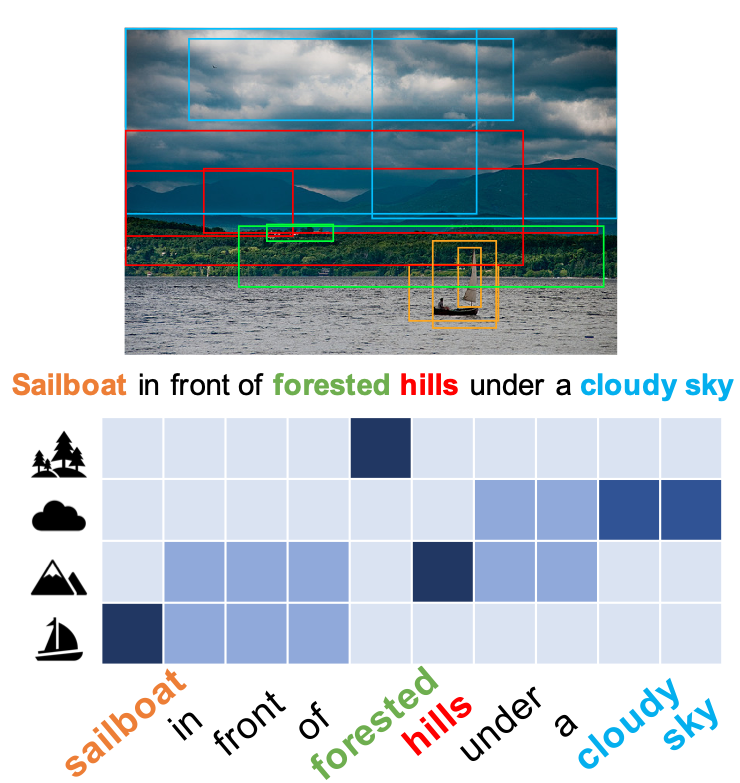}
 \end{subfigure}
 \vspace{-4mm}
    \caption{Illustration of CDA. Left: OT matching scheme for bipartite matching. Strong signals are marked as blue lines. Upper right: manually labeled correspondence; image regions are matched with words of the same color. Lower right: the automatically learned alignment matrix using optimal transport;  darker shades indicate stronger OT matching.}
    \label{fig:intuition}
    \vspace{-6mm}
\end{figure}

The framework developed here makes the following contributions. 
\begin{itemize}
    \vspace{-2.5mm}
    \item Optimal transport is applied to construct principled matching scores for feature alignment across different domains, in particular, images and text.
    \vspace{-2.5mm}
    \item Beyond the functionality as an attention score, OT is also applied as a regularizer on the objective; thus, instead of only being used to match entities within images and text, the proposed OT regularizer can help linking image and text features globally.
    \vspace{-2.5mm}
    \item The effectiveness of our framework is demonstrated on various vision-language tasks, ({\it e.g.}, image-text matching and phrase localization). Experimental results show that the proposed OT-based CDA module provides consistent performance gains on all tasks.
\end{itemize}

\vspace{-4mm}
\section{Background}
\vspace{-2mm}
\paragraph{Optimal transport (OT).} We consider the problem of transporting mass between two discrete distributions supported on some latent feature space $\mathcal{X}$. 
Let $\muv = \{\xv_i, \mu_i\}_{i=1}^n$ and $\nuv = \{ \yv_j, \nu_j \}_{j=1}^m$ be the discrete distributions of interest, where $\xv_i, \yv_j \in \mathcal{X}$ denotes the spatial locations and $\mu_i, \nu_j$, respectively, denote the non-negative masses. Without loss of generality, we assume $\sum_i \mu_i = \sum_j \nu_j = 1$. $\pi \in \BR_+^{n \times m}$ is called a valid transport plan if its row and column marginals match $\muv$ and $\nuv$, respectively, that is to say $\sum_i \pi_{ij} = \nu_j$ and $\sum_j \pi_{ij} = \mu_i$. Intuitively, $\pi$ transports $\pi_{ij}$ units of mass at location $\xv_i$ to new location $\yv_j$. It is known that such transport plans are not unique, and as such, one often seeks a solution $\pi^* \in \Pi(\muv,\nuv)$ that is most preferable in other ways, where $\Pi(\muv,\nuv)$ denotes the set of all viable transport plans. OT finds a solution that is most cost effective w.r.t. some function $C(\xv,\yv)$, in the sense that~\cite{peyre2017computational}
\begin{equation}
\label{eq:ot}
 \begin{array}{rcl} 
    \mathcal{D}(\muv,\nuv)  = 
    \sum_{ij} \pi_{ij}^* C(\xv_i,\yv_j) 
     =  \inf_{\pi \in \Pi(\mu,\nu)} \sum_{ij} \pi_{ij} C(\xv_i,\yv_j) \,,
    \end{array}
\end{equation} 

where $\mathcal{D}(\muv,\nuv)$ is known as the {\it optimal transport distance}.
Hence, $\mathcal{D}(\muv,\nuv)$ minimizes the transport cost from $\muv$ to $\nuv$ w.r.t. $C(\xv,\yv)$.
Of particular interest is the case for which $C(\xv,\yv)$ defines a distance metric on $\CX$, and then $\Dcal(\muv,\nuv)$ induces a distance metric on the space of probability distributions supported on $\CX$, commonly known as the Wasserstein distance \cite{villani2008optimal}. The use of OT allows the flexibility to choose task-specific costs for optimal performance, with examples of Euclidean cost $\| \xv - \yv \|_2^2$ for general probabilistic learning \cite{gulrajani2017improved} and cosine similarity cost $\cos(\xv,\yv)$ for semantic matching tasks \cite{chen2018adversarial}. 
\vspace{-3mm}

\paragraph{Image representation.}
We represent an image as a collection (bag) of feature vectors $\Vmat = \{\vv_k\}_{k=1}^K$, where each $\vv_k \in \R^d$ represents an image entity in feature space, and $K$ is the number of entities.
To simplify our discussion, we identify each entity as a region of interest (RoI), {\it i.e.}, a bounding box, hereafter referred to as a region.
We seek for these features to encode diverse visual concepts, {\it e.g.}, object class, attributes, {\it etc}.

To this end, we follow~\cite{anderson2018bottom}, where $\Fmat = \{\fv_k\}_{k=1}^K$, $\fv \in \R^{2048}$ is obtained from a pre-trained ResNet-101 \cite{he2016deepresidual} concatenated to {\it faster R-CNN}~\cite{ren2015faster} (fR-CNN) on the heavily annotated Visual Genome dataset~\cite{krishna2017visual}. fR-CNN first employs a region proposal network with non-maximum suppression~\cite{neubeck2006efficient} mechanism to propose image regions, then leverages RoI pooling to construct a $2048$-dimensional image feature representation, which is then used for object classification. 
To project the image features into a feature space shared by sentence features (discussed below), we further apply an affine transformation to $\fv_k$:
\vspace{-2mm}
\begin{equation}
    \vv_k = \Wmat_v \fv_k + \bv_v,
    \vspace{-2mm}
\end{equation}
where $\Wmat_v \in \R^{d \times 2048}$ and $\bv_v \in \R^d$ are learnable parameters. 

\vspace{-6mm}
\paragraph{Text sequence representation.}
We follow the setup in \cite{SCAN} to extract feature vectors from the text sequences. Every word (token) is first embedded as a feature vector, and we apply a bi-directional Gated Recurrent Unit (Bi-GRU) \cite{schuster1997bidirectional,bahdanau2014neural} to account for context.
Specifically, let $\Smat = \{\wv_1,...,\wv_M\}$ be a text sequence, where $M$ is the sequence length and $\wv_m$ denotes the $p$-dimensional word embedding vector for the $m$-th word in the sequence.
Then the $m$-th feature vector $\ev_m$ is constructed by averaging the left and right context of the GRU embedding, {\it i.e}. $\ev_m = {\left(\overrightarrow{\hv_m} + \overleftarrow{\hv_m}\right)}/{2}$, where $\overrightarrow{\hv_m} = \overrightarrow{\mbox{GRU}}(\wv_m)$, $\overleftarrow{\hv_m} = \overleftarrow{\mbox{GRU}}(\wv_m)$, here  $\overrightarrow{\hv_m},\overleftarrow{\hv_m},\ev_m\in \sR^d$. Similar to the image features discussed in the last section, we collectively denote these text sequence features as $\mathbf{E} = \{\ev_m\}_{m=1}^M$.

\vspace{-5mm}
\section{Cross-Domain Feature Alignment with OT}\label{sec: OT}
\vspace{-2.5mm}
To motivate our model, we first review some of the favorable properties of Optimal Transport (OT) that appeal to CDA applications. 
\begin{itemize}
\vspace{-2mm}
    \item {\it Sparsity.} It is well known that when solved exactly, OT yields a sparse solution of transportation plan $\pi^*$~\cite{brualdi1991combinatorial}, which eliminates matching ambiguity and facilitates model interpretation ~\cite{de2011optimal}.
    \vspace{-2mm}
    \item {\it Mass conservation.} The solution is self-normalized in the sense that $\pi^*$'s row-sum and column-sum match the desired marginals ~\cite{cuturi2017computational}.
    \vspace{-2mm}
    \item {\it Efficient computation.} OT solutions can be readily approximated using iterative procedures known as Sinkhorn iterations, requiring only matrix-vector products ~\cite{cuturi2013sinkhorn, xie2018fast}.
\end{itemize}
\vspace{-2mm}
Contextualized in a CDA setup, we can regard image and text sequence embeddings as two discrete distributions supported on the same feature representation space. Solving an OT transport plan between the two naturally constitutes a matching scheme to relate cross-domain entities. Alternatively, this allows OT-matching to be viewed as an attention mechanism, as the model attends to the units with high transportation pairing. The OT distance can further serve as a proxy for assessing the global ``relatedness'' between the image and text sequence, {\it i.e.}, a summary of the degree to which the image and text are aligned, justifying its use as a principled regularizer to be incorporated into the training objective. 

To evaluate the OT distance, we first choose a pairwise similarity between $\Vmat$ and $\mathbf{E}$ using a cost function $C(\cdot,\cdot)$.
In our setup, we choose cosine distance  $\Cmat_{km}=C(\ev_k, \vv_m) = 1 - \frac{\ev_k^T \vv_k}{\|\ev_k\|\|\vv_m\|}$ as our cost,
so that (\ref{eq:ot}) can be reformulated as:
\vspace{-3mm}
\begin{align}
    \label{eq:ot-align}
   & \mathcal{L}_{\text{OT}} (\Vmat, \mathbf{E}) = \min_{\Tmat}  \ \sum^K_{k=1}\sum^M_{m=1}\Tmat_{km} \Cmat_{km} 
\end{align} 
   where $\sum_m \Tmat_{km} = \mu_k$, $\,\sum_k \Tmat_{km} = \nu_m$, $\forall k \in [1,K]$, $m \in [1,M]$.
Here, $\Tmat\in \BR_+^{K \times M}$ is the transport matrix, $d_k$ and $d_m$ are the weight of $\vv_k$ and $\ev_m$ in a given image and text sequence, respectively. 
We assume the weight for different features to be uniform, {\it i.e.}, $\mu_k = \frac{1}{K}$, $\nu_m = \frac{1}{M}$.
We leverage the inexact proximal point method optimal transport algorithm (IPOT)~\cite{xie2018fast} to efficiently solve the linear program (\ref{eq:ot-align}).
More details including pseudo-code implementation are summarized in the Supplementary Material (Supp). Below we elaborate on the use of OT-based image-text cross domain alignment in three tasks.

\begin{figure*}[t!]
\scriptsize
  \begin{center}
    \includegraphics[width=0.82\textwidth]{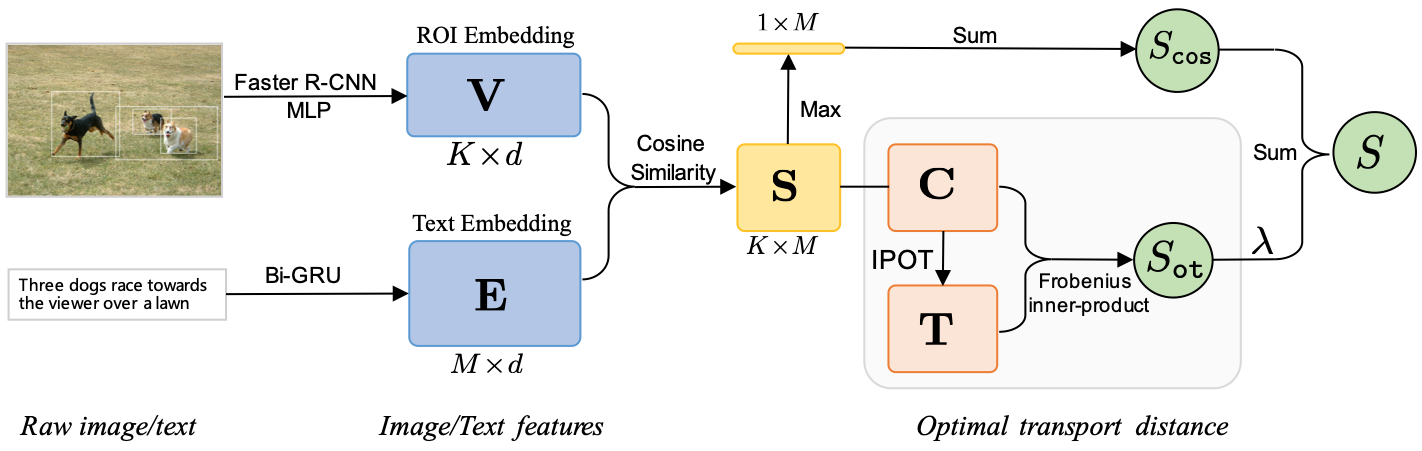}    \vspace{-2mm}
    \caption{Illustration of the proposed retrieval model. Image and text sequences features are represented as bag of feature vectors (in blue).  Cosine similarity matrix $\Smat$ is computed (in yellow). Two types of similarity measures are considered: (1) the traditional sum-max text-image aggregation $S_{\cos}$ and  (2) optimal transport $S_{\text{OT}}$ (in green circle). The final score $S$ is obtained as the weighted sum of the two similarity scores.}
    \label{fig:model_retrieval}
  \end{center}
  \vspace{-5mm}
\end{figure*}

\vspace{-6mm}
\paragraph{Image-Text Matching.}
\label{sec: model}
We start our discussion with image-text matching, a building block of cross-modal retrieval tasks required by many downstream applications. 
In image-text matching, a model searches for a matching image in an image library based on a text description, or searches for a matching caption in a caption library based on an image.  Figure~\ref{fig:model_retrieval} presents a diagram of the proposed OT-based CDA.
The feature vectors $\Vmat$ and $\Emat$ are extracted from images and text sequences using the fR-CNN and Bi-GRU models, respectively. 
The similarity between an image and a text sequence is obtained in terms of two types of similarity measures, computed over all possible entities (regions and words) within an image and text sequence.
Specifically, we consider $i$) a sum-max text-image  aggregated cosine similarity, and $ii$) a weighted OT-based similarity that explicitly accounts for all similarities between pairs of entities, as detailed below.

{\it Baseline similarity score}:
Following the practice of \cite{karpathy2015deep}, we first derive a global similarity score as the baseline target to optimize for cross-domain alignment. 
More specifically, we begin by computing the pairwise similarities for the $k$-th region and the $m$-th token using {\it cosine similarity} in the feature space
\vspace{-1mm}
\begin{equation}
\label{eq:dot_sim}
    s_{km} =\frac{\vv_k^\top {\ev}_m}{\|\vv_k\| \|{\ev}_m\|}, \ \ k \in [1, K], \ \ m \in [1, M]\,.
    \vspace{-1mm}
\end{equation}
The {\it global similarity score} is built via the aggregation: 
\vspace{-3mm}
\begin{equation}\label{eq:sum_max_ti}
    S_{\cos}(\Vmat,\Emat) = \sum_{m=1}^M\max_{k}(s_{km}).
    \vspace{-3mm}
\end{equation}
This strategy is known as \textit{sum-max text-image} aggregation and has been applied successfully in image-text matching tasks.  
Alternatively, we can use \textit{sum-max image-text}, where $S_{\cos}'(\Vmat,\Emat) = \sum_{k=1}^K\max_{m}(s_{km})$.
Our choice of {\it sum-max text-image} is based on the ablation study from \cite{SCAN} and \cite{harwath2018jointly}, which showed empirical evidence that \textit{sum-max text-image} works better in practice.

{\it OT similarity score:} In addition to the above {\it sum-max text-image} score, we present an OT construction of global similarity, which is the key regularizer in our framework.
Specifically, we choose the cost matrix $\Cmat$ to be $\Cmat_{km} = 1-s_{km}$.
The OT-based similarity score can be defined as $S_{\text{OT}}(\Vmat, \Emat) = -\mathcal{L}_{\text{OT}}(\Vmat, \Emat)$ using (\ref{eq:ot-align}), 
and the transport plan $\Tmat$ naturally corresponds to the  cross-domain alignment strategy.

{\it Composed similarity score:} We integrate both similarity scores discussed above using a simple linear combination
\vspace{-3mm}
\begin{equation}\label{eq: sim_score}
    S(\Vmat, \Emat) = S_{\cos}(\Vmat, \Emat) + \lambda S_{\text{OT}}(\Vmat, \Emat)\, ,
    \vspace{-3mm}
\end{equation}
where $\lambda$ is a hyper-parameter weighting the relative importance of the OT similarity score.
Intuitively, the baseline $S_{\cos} (\Vmat, \Emat)$ provides an {\it unweighted} account of the aggregated agreement between regions and words of an image and text sequence,
while the OT-based $S_\text{OT}$ provides a {\it weighted} summary of how well every region in the image matches every word in the text sequence.
Consequently, (\ref{eq: sim_score}) accounts for both aggregated and weighted alignments to assess global similarity.
From the perspective of attention models, $S_{\cos}$ can be understood as a {\it hard} attention whereas $S_\text{OT}$ can be perceived as a {\it soft} attention.
The hyperparameter $\lambda$ controls the smoothness and sparsity of the final matching plan, which gives the similarity score S(\Vmat, \Emat).

{\it Final training objective}: To derive our final training loss, we consider the construction known as {\it triplet loss with hardest negatives}, originally proposed in \cite{wang2016learning,VSE}.
For each batch of $B$ image and sentence pairs $\{\Vmat_j, \Emat_j\}_{j=1}^B$, the total loss is given by
%
\vspace{-3mm}
\begin{equation}
\label{eq:objective}
    \Lcal =  \sum^B_{j=1} \biggl\{\max\left[0, S\left(\Vmat_j, \Emat_j^-\right)-S\left(\Vmat_j, \Emat_j\right)+\eta\right]  
     + \max\left[0, S\left(\Vmat_j^-, \Emat_j\right)-S\left(\Vmat_j, \Emat_j\right)+\eta\right]\biggr\}\,,
    \vspace{-3mm}
\end{equation}
where the hardest negatives are given by $\Vmat_j^- = \argmax_{\vv \in \Vmat_{\backslash j}}S(\vv, \Emat_j)$ and \\$\Emat_j^- = \argmax_{\ev \in \Emat_{\backslash j}}S(\Vmat_j, \ev)$, and $\{\backslash j\}$ denotes all indices except for $j$. 

This means that once the score of the positive pair, \ie $S\left(\Vmat_j, \Emat_j\right)$, is higher by $\eta$ units over the score for the negative pair with the highest score in a batch, the hinge loss is zero.
This training objective encourages separation in similarity score between paired data and unpaired data.
\begin{table}[h!]\footnotesize
  \begin{center}
  \caption{
Cross-domain matching results with Recall@$K$ (R@K). Upper panel: Flickr30K, lower panel: MSCOCO. 
  }
    \vspace{-0mm}
  \label{table:retrieval}
  \centering
    \begin{tabular*}{1.\linewidth}{@{\extracolsep{\fill}}lccccccc}
         \hline
         \small
         & \multicolumn{3}{c}{Sentence Retrieval} & \multicolumn{3}{c}{Image Retrieval} & \multicolumn{1}{c}{}\\
        Method & R@1 & R@5 & R@10 & R@1 & R@5 & R@10 &Rsum\\
        \hline
        DVSA (R-CNN, AlexNet) \cite{karpathy2015deep} & 22.2 & 48.2 & 61.4 & 15.2 & 37.7 & 50.5 & 235.2\\ 
        HM-LSTM (R-CNN, AlexNet) \cite{niu2017hierarchical} & 38.1 & -- & 76.5 & 27.7 & -- & 68.8 & --\\ 
        2WayNet (VGG) \cite{eisenschtat2017linking} & 49.8 & 67.5 & -- & 36.0 & 55.6 & -- & --\\ 
        SM-LSTM (VGG) \cite{huang2017instance} & 42.5 & 71.9 & 81.5 & 30.2 & 60.4 & 72.3 & 358.8\\ 
        VSE++ (ResNet) \cite{VSE} & 52.9 & -- & 87.2 & 39.6 & -- & 79.5 & -- \\ 
        DPC (ResNet) \cite{zheng2017dual} & 55.6 & 81.9 & 89.5 & 39.1 & 69.2 & 80.9 & 416.2\\ 
        DAN (ResNet) \cite{nam2017dual} & 55.0 & 81.8 & 89.0 & 39.4 & 69.2 & 79.1 & 413.5\\ 
        SCO (ResNet) \cite{huang2018learning} & 55.5 & 82.0 & 89.3 & 41.1 & 70.5 & 80.1 & 418.5\\ 
        SCAN (Faster R-CNN, ResNet) \cite{SCAN}  & 67.7 & 88.9 & 94.0 & 44.0 & 74.2 & 82.6 &452.2 \\
        BFAN (Faster R-CNN, ResNet)\cite{liu2019focus} & 65.5 & 89.4 & - & 47.9 & 77.6 & - & - \\ 
        PFAN (Faster R-CNN, ResNet)\cite{wang2019position} & 66 & 89.6 & 94.3 & 49.6 & 77 & 84.2 & 460.7 \\
        VSRN (Faster R-CNN, ResNet)\cite{li2019visual}  & 65 & 89 & 93.1 & 49 & 76 & 84.4 & 456.5\\
        \textbf{Ours (Faster R-CNN, ResNet)}:\\
        $\cos$ + OT & \textbf{69} & \textbf{91.8} & \textbf{95.9} & \textbf{50.4} & \textbf{77.6} & \textbf{85.5} & \textbf{470.2} \\
        \hline
        \noalign{\smallskip}
        Order-embeddings (VGG) \cite{vendrov2015order} & 23.3 & -- & 84.7 & 31.7 & -- & 74.6 & -- \\ 
        VSE++ (ResNet) \cite{VSE} & 41.3 & -- & 81.2 & 30.3 & -- & 72.4 & -- \\ 
        DPC (ResNet) \cite{zheng2017dual} & 41.2 & 70.5 & 81.1 & 25.3 & 53.4 & 66.4 & 337.9\\ 
        GXN (ResNet) \cite{gu2018look} & 42.0 & -- & 84.7 & 31.7 & -- & 74.6 & -- \\ 
        SCO (ResNet) \cite{huang2018learning} & 42.8 & 72.3 & 83.0 & 33.1 & 62.9 & 75.5 & 369.6 \\ 
        SCAN (Faster R-CNN, ResNet)\cite{SCAN} & 46.4 & 77.4 & 87.2 & 34.4 & 63.7 & 75.7 & 384.8 \\ 
        VSRN (Faster R-CNN, ResNet)\cite{li2019visual}  & 48.6 & 78.9 & 87.7 & \textbf{37.8} & \textbf{68} & {77.1} & 398.1 \\
        \textbf{Ours (Faster R-CNN, ResNet)}:\\
        $\cos$ + OT &\textbf{49.9} &\textbf{81.4} &\textbf{89.8} &\textbf{37.8} &{66.7} &\textbf{78.1} &\textbf{403.6} \\
        \hline
    \end{tabular*}

    \vspace{-5mm}
  \end{center}
\end{table}

\vspace{-4mm}
\begin{figure*}[t!]
  \begin{center}
    \includegraphics[width=0.8\textwidth]{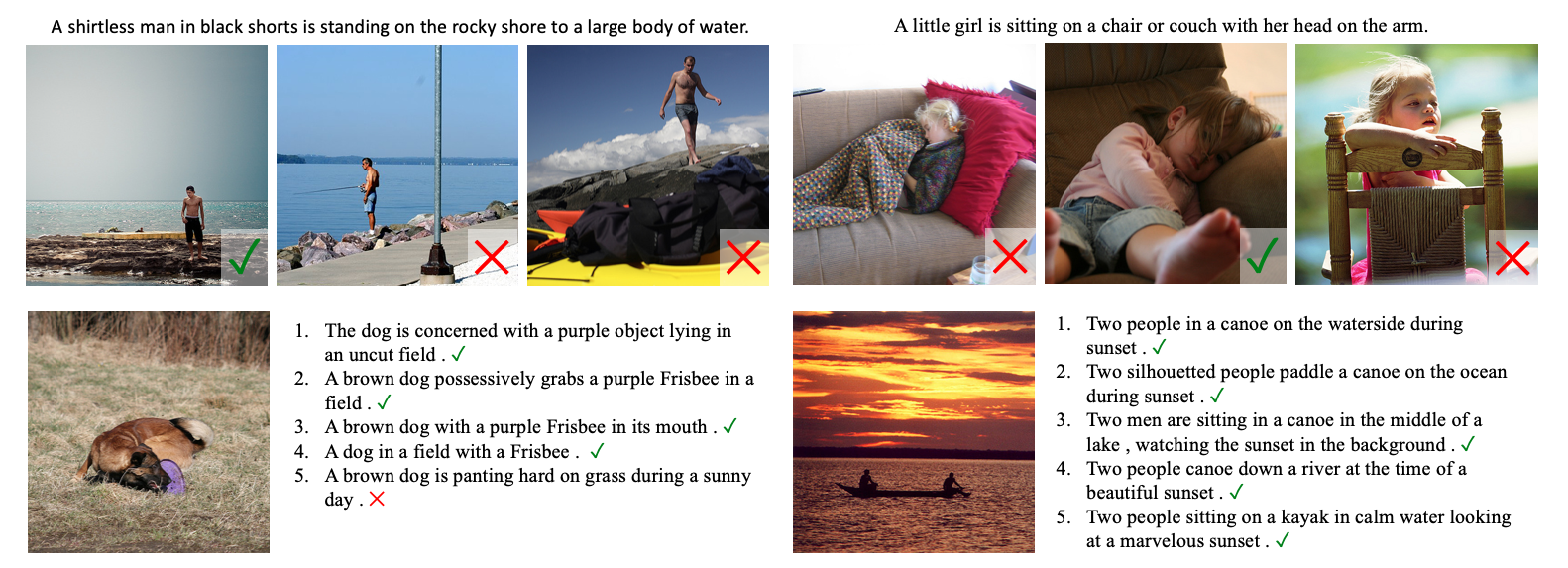}
    \vspace{-2mm}
    \caption{\small Examples of image-text retrieval results using OT regularization. First row shows text-to-image retrieval results. For each sentence, the top-3 matched images are listed from left to right. Right-bottom corner in each image indicates if this is a ground truth image. Image-to-text retrieval results are shown in the second row, where the top-5 sentences given an image query are provided. The mark at the end of each sentence denotes if this is a ground truth sentence. Throughout the text a green checks indicates ground-truth, while a red cross indicates otherwise. 
    }
    \label{fig:t2i_sample}
    \vspace{-2mm}
  \end{center}
\end{figure*}
\vspace{-2mm}
\paragraph{Weakly supervised phrase localization.}
The phrase-localization task aims to learn relatedness between text phrases and image regions. Weakly supervised phrase localization guided by an image-sentence pair can serve as an evaluation of CDA methods, as the performance of phrase localization shows the model's ability to capture vision-language interactions. Phrase localization seeks to learn a mapping model $f(k|m,\Imat,\wv)$ that evaluates the probability that the $m$-th token in text sequence $\wv$ references the $k$-th region in image $\Imat$. For our model, we define the mapping model as:
\vspace{-2mm}
\begin{equation}
\label{eq:M=T}
    \vspace{-2mm}
    f(k|m,\Imat,\wv) \propto \Tmat_{km}.
    \vspace{-2mm}
\end{equation}
For the baseline model, we use the cosine similarity matrix as the mapping model:
\vspace{-2mm}
\begin{equation}
\label{eq:M=S}
    f(k|m,\Imat,\wv) \propto s_{km}.
    \vspace{-2mm}
\end{equation}
For each model, we first train with the image-sentence matching task, and then directly apply the model to the phrase localization task without further tuning.


\vspace{-3mm}
\section{Related Work}
\vspace{-2mm}
\paragraph{Optimal transport.}
Efforts have been made to use OT to find intra-domain similarities.
In computer vision, the earth mover's distance (EMD), also known as the OT distance, is used to match the distribution of the content between two images~\cite{rubner2000earth}.
OT has also been applied successfully to NLP tasks such as document classification ~\cite{kusner2015word}, sequence-to-sequence learning~\cite{chen2019improving} and text generation~\cite{chen2018adversarial}.
In these works, OT has been applied to within-domain alignment, either for image regions or text sequences, capturing the intra-domain semantics.
This paper constitutes the first work to use OT for cross-domain feature alignment, {\it e.g.} in image-text retrieval and weakly supervised phrase grounding tasks.
\vspace{-6mm}

\paragraph{Image-text matching.}
Many works have investigated embedding image and text sequence features into a joint semantic space for image-text matching.
The first attempt was made by~\cite{kiros2014unifying}, where the authors proposed to use CNNs to encode the images and LSTMs to encode text.
The model was trained with a hinge-base triplet ranking loss, and 
it was improved by adding hardest negatives in the triplet ranking loss~\cite{VSE}.
To consider the relationship between image regions and text sequences, \cite{karpathy2015deep} first computed the similarity matrix for all regions and word pairs via a dot product, and then calculated the similarity score with a sum or max aggregation function (denoted as \textit{dot} model). 
Recently, SCAN~\cite{SCAN} was proposed to use two-step stacked cross attention to measure similarities in image region and text pairs. Further works include VSRN\cite{li2019visual}, PFAN\cite{wang2019position} and BFAN\cite{liu2019focus}. 
In our model, we share the same motivation as SCAN, but we propose OT to obtain the optimal relevance correspondence between entities from the two domains. 
Recently, large-scale vision-language pre-training~\cite{li2020oscar,sun2019videobert,tan2019lxmert,lu2019vilbert,chen2019uniter,su2019vl,li2019unicoder,hao2020prevalent} has provided more informative representations for image-text pairs, and has achieved state-of-the art matching performance. The proposed OT is orthogonal to this, and we leave it as future research to combine OT with pre-trained models.
\vspace{-2mm}

\vspace{-2mm}
\paragraph{Weakly supervised phrase localization.}
Motivated by the large amount of annotation efforts for supervised approaches, some previous works~\cite{Align2ground, zhao2018weakly, chen2018knowledge, Engilberge_2018_CVPR, karpathy2015deep} attempted to use matched image-text pairs as supervision to guide phrase localization training. In \cite{karpathy2015deep, Align2ground} a local region-phrase similarity score was calculated, followed by  aggregating the local scores to calculate the global image-text similarity score.

\vspace{-2mm}
\section{Experiments}
\vspace{-2mm}
\paragraph{Datasets.}
We evaluate our model on the Flickr30K \cite{plummer2015flickr30k} and MS-COCO \cite{lin2014MSCOCO} datasets. Flickr30K contains $31$,$000$ images, and each photo has five human-annotated captions.
We split the data following the same setup as \cite{karpathy2015deep,VSE}.
The dataset has $29$,$000$ training images, $1$,$000$ validation images and $1$,$000$ test images.
MS-COCO contains $123$,$287$ images, and each image is annotated with $5$ human-generated text descriptions. 
We use the split in \cite{VSE}, {\it i.e.}, the dataset is split into $113$,$287$ training images, $5$,$000$ validation images and $5$,$000$ test images.

\vspace{-2mm}
\begin{figure*}[h!]
    \vspace{-3.5mm}
    \centering
        \includegraphics[width=1.\textwidth,trim={1cm 15.5cm 1cm 3cm}, clip]{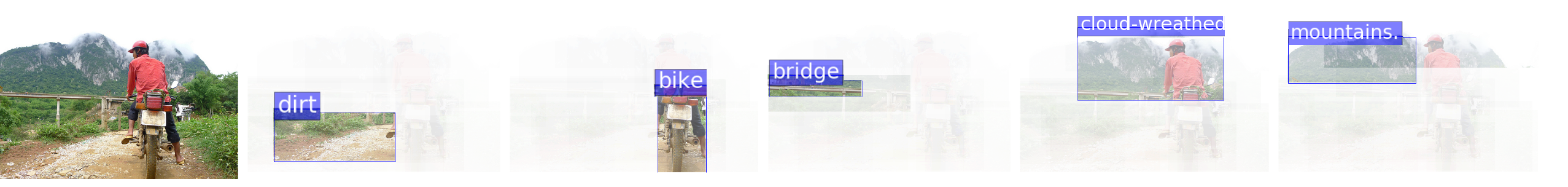}
        \vspace{-8mm}
        \caption{\small Visualization of the learned OT alignment. We show attended image regions with matched key words. The brightness reflects the alignment strength. 
        The left-most figure is the original image. 
        Each bounding box is the region with highest OT alignment score w.r.t the matched key word. Our model successfully identifies the correct pairing without seeing any ground-truth ({\it i.e.}, weak supervision) during training. 
        }
      \label{fig:VisAlignment}
    \vspace{-4.5mm}
    \end{figure*}
\begin{figure*}[h!]
\centering
    \includegraphics[width=0.75\textwidth,trim={1cm 9cm 8.5cm 5.3cm}, clip]{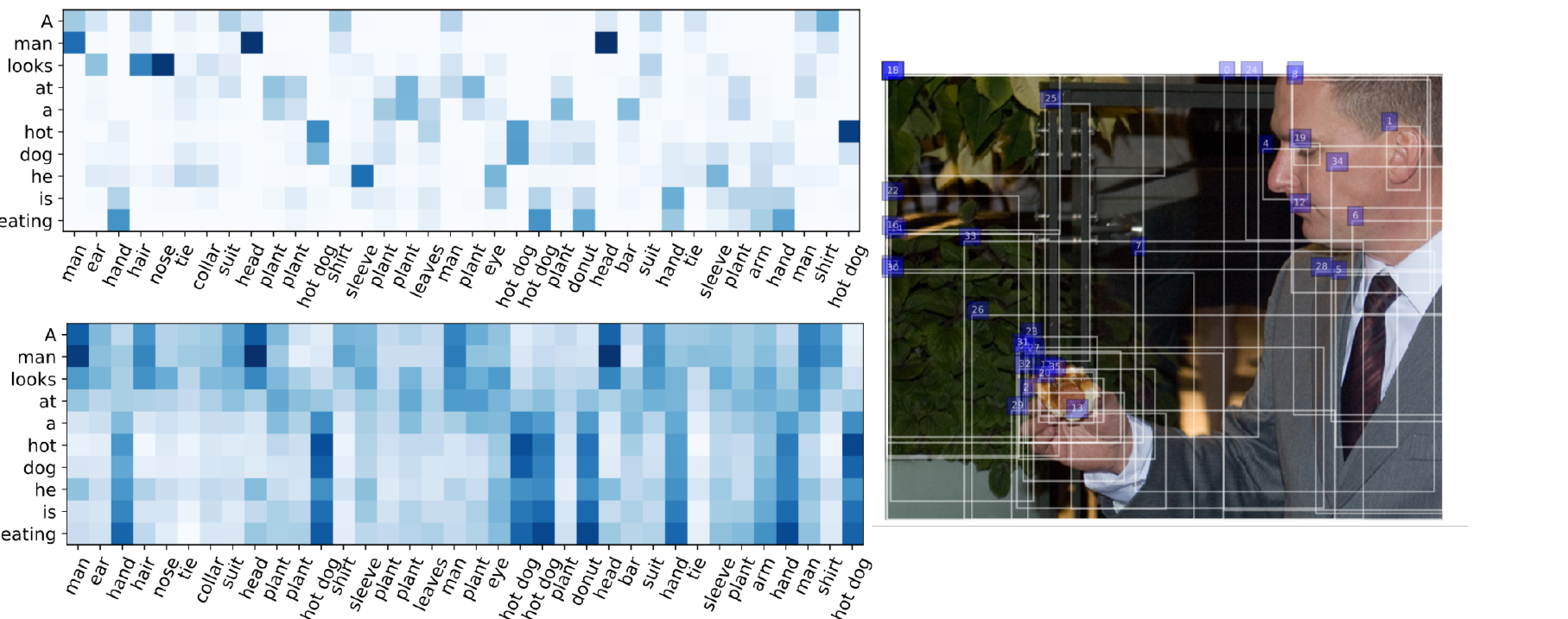}
    \vspace{-4mm}
    \caption{\small A comparison of OT transport matrix (top left) and attention matrix (bottom left). The horizontal axis represents image regions (annotated here to facilitate understanding), and the vertical axis represents words. Original image on the right. }
  \label{fig:OTvsAttention}
\vspace{-6mm}
\end{figure*}
\paragraph{Implementation details.}
For image-text matching, we use the Adam optimizer~\cite{kingma2014adam} to train the models.
For the Flickr30K data, we train the model for 30 epochs.
The initial learning rate is set to 0.0002, and decays by a factor of 10 after 15 epochs.
For MS-COCO data, we train the model for 20 epochs.
The initial learning rate is set to $0.0005$, and decays by 10 after 10 epochs.
The batch size to 128, and the maximum gradient norm is thresholded to 2.0 for gradient clipping.
The dimension of the GRU and joint embedding space is set to $1800$, and the dimension of the word embedding to 300.
Twenty iterations of the IPOT algorithm are considered. Since the performance of a single model is not reported in VSRN \cite{li2019visual} paper, we ran associated experiments based on their github repository.

\vspace{-2mm}
\subsection{Image-text matching}
\vspace{-2mm}
We evaluate image-text matching on both datasets.
The performance of sentence retrieval with image query or image retrieval with sentence query is measured by recall at $K$ (R@K)~\cite{karpathy2015deep}, defined as the percentage of queries which retrieve the correct object within those with top $K$ highest similarity scores as determined by the model.
For each retrieval task, $K=\{1,5,10\}$ is recorded.
We use Rsum~\cite{huang2017instance} to evaluate the overall performance, defined as: $\text{Rsum} = \sum_K \text{R@K}_{\text{I2T}} + \text{R@K}_{\text{T2I}}$, where I2T denotes image-to-text retrieval and T2I denotes text-to-image retrieval.

Table~\ref{table:retrieval} shows the quantitative results on Flickr30K and MS-COCO, with $\eta$ representing the margin in (\ref{eq:objective}) and $\lambda$ the weight on the OT regularizer in (\ref{eq: sim_score}).
Hyper-parameters $\eta$ and $\lambda$ are determined with a grid search using the validation set, specifically, $\eta=0.12$, $\lambda=1.5$ for Flickr30K, and $\eta=0.05$, $\lambda=0.1$ for MS-COCO.
We see that for a single model, our approach outperforms or is comparable with the current state-of-the-art method VSRN \cite{li2019visual}. Similar results are observed under an ensemble setup (see Supp for detailed results).

\vspace{-2mm}
\subsection{Weakly supervised phrase localization}
\vspace{-2mm}
In order to demonstrate the efficiency of our CDA method under weak-supervision, we executed the weakly supervised phrase grounding experiment using pretrained retrieval models described in Section \ref{sec: OT}. Our implementation is based on Bilinear Attention Network codebase\footnote{\footnotesize \url{https://github.com/jnhwkim/ban-vqa}}.  We evaluate the models with the percentage of phrases that are correctly localized with respect to the ground truth bounding box across all images, where correct localization is defined as IoU $\geq 0.5$~\cite{plummer2015flickr30k}. Specifically, $K$ predictions are permitted to find at least one correction, called Recall at $K$ (R@K). Table \ref{table:grounding} shows the comparison between our model and baseline SCAN model on Flickr30k~\cite{plummer2015flickr30k}. When training the retrieval model, we choose the set of hyper-parameter that achieves best performance for both our model and the baseline SCAN model. In particular, {\it OT\_T} represents the model described in Eq. (\ref{eq:M=T}), and {\it OT\_S} represents the model described in Eq. (\ref{eq:M=S}) with image/text encoders trained by our model. Our approach outperforms the baseline model on all three metrics. This indicates that by leveraging OT, not only better alignment is computed, but also better feature encoders are trained.
\begin{table}[h!]\footnotesize
    \begin{center}
    \caption{
  Phrase localization result on Flickr30K Entities
    }
    \label{table:grounding}
    \centering
        \begin{tabular}{lccc}
           \toprule
          Method & R@1 & R@5 & R@10\\
          \toprule
          SCAN & 20.79 & 47.45 & 55.14 \\
        Dot & 35.09 & 64.35 & 68.48 \\
        MATN \cite{zhao2018weakly} & 33.10 & - & - \\
        KAC Net\cite{chen2018knowledge} & 38.71 & - & - \\
        \hline
        OT\_T & 35.98 & 70.33 & 78.97 \\
        OT\_S & 41.12 & 70.42 & 77.48 \\
            \bottomrule
      \end{tabular}
  
      \vspace{-5mm}
    \end{center}
  \end{table}

\begin{table*}[h!] \footnotesize
    \begin{center}
        \caption{ Ablation study on Flickr30K. We study the impact of hyper-parameters for OT and the baseline.}
        \label{table:ablation}
            \centering
        \begin{tabular*}{1.\linewidth}{@{\extracolsep{\fill}}lccccccc}
            \toprule
             & \multicolumn{3}{c}{Sentence Retrieval} & \multicolumn{3}{c}{Image Retrieval} & \multicolumn{1}{c}{} \\
            Method & R@1 & R@5 & R@10 & R@1 & R@5 & R@10 & Rsum\\
            \hline
            \noalign{\smallskip}
            $\cos$, $\eta$=0.2  & 61.7 & 87.4 & 93.5 & 48.5 & 76.0 & 83.7 & 450.8 \\ 
            $\cos$ + OT, $\eta$=0.2, $\lambda$=1 & 66.2 & 89.0 & 94.1 & 48.9 & 77.5 & {85.4} &461.1 \\ 
            $\cos$, $\eta$=0.12  & 63.1 &89.5 & 94.3 & {50.5} & 77.1 & 84.7&459.2 \\ 
            $\cos$ + OT, $\eta$=0.12, $\lambda$=2 & {69.3} & {91.0} &{95.7} & 48.4 & {77.2} & 84.7 &{466.3} \\
            \bottomrule
        \end{tabular*}
    \end{center}
\end{table*}

\vspace{-2mm}
\subsection{Qualitative results}
\vspace{-2mm}
We provide samples of image-text retrieval results from Flickr30K test set in Figure \ref{fig:t2i_sample}.
For each sentence query, we present the top-3 images ranked by similarity score, as calculated by our model.
For each image query, we present the top-5 sentences.
From this representative sample we see that our model matches images and sentences with high correlation.
Although the query text and retrieved images (and query image and retrieved text) are not the exact pairs, they are still highly correlated and share the same theme. More qualitative results for image-text retrieval, image captioning and VQA are presented in Supp.

\vspace{-2mm}
\subsection{Analysis}
\vspace{-1mm}
\textbf{Ablation study.}
We consider several ablation settings to further examine the capabilities of the proposed OT algorithm.
To show the effectiveness of OT, we consider an ablation experiment for the image-text retrieval task. 
In Table \ref{table:ablation}, we compare our model with the baseline, which only uses cosine similarity to measure the distance between image and text features, {\it i.e.}, only (\ref{eq:sum_max_ti}) is applied.
Two hyper-parameter combinations are considered.
In both cases, the OT-enhanced similarity outperforms the baseline model, demonstrating the effectiveness of optimal transport. The ablation study on network architecture choices and adaptive region numbers are found in the Supp.

\noindent\textbf{Interpretable alignment.} 
One favorable property of OT is the interpretability of the optimal transport plan $\Tmat$. 
To illustrate this, we visualize $\Tmat$ in comparison with the attention matrix $(1-\Cmat)$ in Figure \ref{fig:OTvsAttention}. The darker shade implies stronger OT matching or attention weights.
We see that OT transport mapping is more interpretable, as the alignment is sparse and self-normalized. See Supp for more examples.

\vspace{-3mm}
\section{Conclusions}
\vspace{-2.5mm}
We have proposed to use optimal transport to provide a principled alignment between features from text-image domains.
We take advantage of such alignment when computing similarity scores for image and text entities in matching tasks, and the results outperform the state of the art. 
Moreover, we show the accuracy of OT-based alignment with phrase localization and achieve better performance than baseline models.
As future work, it is of interest to take advantage of OT alignment in other text-image cross-domain tasks, such as visual question answering and text-to-image generation. 
\section{Acknowledgements}
The authors would like to thank the anonymous reviewers for their insightful comments. The research at Duke University was supported in part by DARPA, DOE, NIH, NSF and ONR.

\clearpage
{
\bibliography{ref}

\begin{thebibliography}{65}
\providecommand{\natexlab}[1]{#1}
\providecommand{\url}[1]{\texttt{#1}}
\expandafter\ifx\csname urlstyle\endcsname\relax
  \providecommand{\doi}[1]{doi: #1}\else
  \providecommand{\doi}{doi: \begingroup \urlstyle{rm}\Url}\fi

\bibitem[Anderson et~al.(2018)]{anderson2018bottom}
Peter Anderson et~al.
\newblock Bottom-up and top-down attention for image captioning and visual
  question answering.
\newblock In \emph{CVPR}, pages 6077--6086, 2018.

\bibitem[Antol et~al.(2015)]{antol2015vqa}
Stanislaw Antol et~al.
\newblock Vqa: Visual question answering.
\newblock In \emph{ICCV}, pages 2425--2433, 2015.

\bibitem[Arjovsky et~al.(2017)]{arjovsky2017wasserstein}
Martin Arjovsky et~al.
\newblock {W}asserstein generative adversarial networks.
\newblock In \emph{ICML}, 2017.
\newblock URL \url{http://proceedings.mlr.press/v70/arjovsky17a.html}.

\bibitem[Bahdanau et~al.(2015)Bahdanau, Cho, and Bengio]{bahdanau2014neural}
Dzmitry Bahdanau, Kyunghyun Cho, and Yoshua Bengio.
\newblock Neural machine translation by jointly learning to align and
  translate.
\newblock In \emph{ICLR}, 2015.

\bibitem[Brualdi and Ryser(1991)]{brualdi1991combinatorial}
Richard~A Brualdi and Herbert~J Ryser.
\newblock \emph{Combinatorial matrix theory}, volume~39.
\newblock 1991.

\bibitem[Chen et~al.(2018{\natexlab{a}})Chen, Gao, and
  Nevatia]{chen2018knowledge}
Kan Chen, Jiyang Gao, and Ram Nevatia.
\newblock Knowledge aided consistency for weakly supervised phrase grounding.
\newblock In \emph{Proceedings of the IEEE Conference on Computer Vision and
  Pattern Recognition}, pages 4042--4050, 2018{\natexlab{a}}.

\bibitem[Chen et~al.(2019{\natexlab{a}})Chen, Zhang, et~al.]{chen2019improving}
Liqun Chen, Yizhe Zhang, et~al.
\newblock Improving sequence-to-sequence learning via optimal transport.
\newblock In \emph{ICLR}, 2019{\natexlab{a}}.

\bibitem[Chen et~al.(2018{\natexlab{b}})]{chen2018adversarial}
Liqun Chen et~al.
\newblock Adversarial text generation via feature-mover's distance.
\newblock In \emph{{NeurIPS}}, 2018{\natexlab{b}}.

\bibitem[Chen et~al.(2019{\natexlab{b}})Chen, Li, Yu, Kholy, Ahmed, Gan, Cheng,
  and Liu]{chen2019uniter}
Yen-Chun Chen, Linjie Li, Licheng Yu, Ahmed~El Kholy, Faisal Ahmed, Zhe Gan,
  Yu~Cheng, and Jingjing Liu.
\newblock Uniter: Learning universal image-text representations.
\newblock \emph{arXiv preprint arXiv:1909.11740}, 2019{\natexlab{b}}.

\bibitem[Cuturi and Peyr{\'e}(2017)]{cuturi2017computational}
M~Cuturi and G~Peyr{\'e}.
\newblock Computational optimal transport.
\newblock 2017.

\bibitem[Cuturi(2013)]{cuturi2013sinkhorn}
Marco Cuturi.
\newblock Sinkhorn distances: Lightspeed computation of optimal transport.
\newblock In \emph{{NeurIPS}}, pages 2292--2300, 2013.

\bibitem[Datta et~al.(2019)Datta, Sikka, Roy, Ahuja, Parikh, and
  Divakaran]{Align2ground}
Samyak Datta, Karan Sikka, Anirban Roy, Karuna Ahuja, Devi Parikh, and Ajay
  Divakaran.
\newblock Align2ground: Weakly supervised phrase grounding guided by
  image-caption alignment.
\newblock In \emph{ICCV}, 03 2019.

\bibitem[De~Goes et~al.(2011)]{de2011optimal}
Fernando De~Goes et~al.
\newblock An optimal transport approach to robust reconstruction and
  simplification of 2d shapes.
\newblock In \emph{Computer Graphics Forum}, volume~30, 2011.

\bibitem[Eisenschtat and Wolf(2017)]{eisenschtat2017linking}
Aviv Eisenschtat and Lior Wolf.
\newblock Linking image and text with 2-way nets.
\newblock In \emph{CVPR}, 2017.

\bibitem[Engilberge et~al.(2018)Engilberge, Chevallier, Pérez, and
  Cord]{Engilberge_2018_CVPR}
Martin Engilberge, Louis Chevallier, Patrick Pérez, and Matthieu Cord.
\newblock Finding beans in burgers: Deep semantic-visual embedding with
  localization.
\newblock In \emph{CVPR}, June 2018.

\bibitem[Faghri et~al.(2018)Faghri, Fleet, Kiros, and Fidler]{VSE}
Fartash Faghri, David~J Fleet, Jamie~Ryan Kiros, and Sanja Fidler.
\newblock Vse++: Improved visual-semantic embeddings.
\newblock In \emph{BMVC}, volume~2, page~8, 2018.

\bibitem[Fang et~al.(2015)Fang, Gupta, et~al.]{fang2015captions}
Hao Fang, Saurabh Gupta, et~al.
\newblock From captions to visual concepts and back.
\newblock In \emph{CVPR}, pages 1473--1482, 2015.

\bibitem[Gu et~al.(2018)Gu, Cai, Joty, Niu, and Wang]{gu2018look}
Jiuxiang Gu, Jianfei Cai, Shafiq~R Joty, Li~Niu, and Gang Wang.
\newblock Look, imagine and match: Improving textual-visual cross-modal
  retrieval with generative models.
\newblock In \emph{CVPR}, 2018.

\bibitem[Gulrajani et~al.(2017)]{gulrajani2017improved}
Ishaan Gulrajani et~al.
\newblock Improved training of {Wasserstein GANs}.
\newblock In \emph{{NeurIPS}}, 2017.

\bibitem[Hao et~al.(2020)Hao, Li, Li, Carin, and Gao]{hao2020prevalent}
Weituo Hao, Chunyuan Li, Xiujun Li, Lawrence Carin, and Jianfeng Gao.
\newblock Towards learning a generic agent for vision-and-language navigation
  via pre-training.
\newblock In \emph{CVPR}, 2020.

\bibitem[Harwath et~al.(2018)Harwath, Recasens, Sur{\'\i}s, Chuang, Torralba,
  and Glass]{harwath2018jointly}
David Harwath, Adria Recasens, D{\'\i}dac Sur{\'\i}s, Galen Chuang, Antonio
  Torralba, and James Glass.
\newblock Jointly discovering visual objects and spoken words from raw sensory
  input.
\newblock In \emph{ECCV}, 2018.

\bibitem[He et~al.(2016)He, Zhang, Ren, and Sun]{he2016deepresidual}
Kaiming He, Xiangyu Zhang, Shaoqing Ren, and Jian Sun.
\newblock Deep residual learning for image recognition.
\newblock In \emph{CVPR}, pages 770--778, 2016.

\bibitem[Hochreiter et~al.(1997)]{hochreiter1997long}
Sepp Hochreiter et~al.
\newblock Long short-term memory.
\newblock \emph{Neural computation}, 1997.

\bibitem[Huang et~al.(2017)Huang, Wang, and Wang]{huang2017instance}
Yan Huang, Wei Wang, and Liang Wang.
\newblock Instance-aware image and sentence matching with selective multimodal
  lstm.
\newblock In \emph{CVPR}, pages 2310--2318, 2017.

\bibitem[Huang et~al.(2018)Huang, Wu, Song, and Wang]{huang2018learning}
Yan Huang, Qi~Wu, Chunfeng Song, and Liang Wang.
\newblock Learning semantic concepts and order for image and sentence matching.
\newblock In \emph{CVPR}, pages 6163--6171, 2018.

\bibitem[Johnson et~al.(2016)Johnson, Karpathy, and
  Fei-Fei]{johnson2016densecap}
Justin Johnson, Andrej Karpathy, and Li~Fei-Fei.
\newblock Densecap: Fully convolutional localization networks for dense
  captioning.
\newblock In \emph{CVPR}, pages 4565--4574, 2016.

\bibitem[Karpathy and Fei-Fei(2015)]{karpathy2015deep}
Andrej Karpathy and Li~Fei-Fei.
\newblock Deep visual-semantic alignments for generating image descriptions.
\newblock In \emph{CVPR}, pages 3128--3137, 2015.

\bibitem[Kingma and Ba(2015)]{kingma2014adam}
Diederik~P Kingma and Jimmy Ba.
\newblock Adam: A method for stochastic optimization.
\newblock \emph{ICLR}, 2015.

\bibitem[Kiros et~al.(2014)Kiros, Salakhutdinov, and Zemel]{kiros2014unifying}
Ryan Kiros, Ruslan Salakhutdinov, and Richard~S Zemel.
\newblock Unifying visual-semantic embeddings with multimodal neural language
  models.
\newblock In \emph{{NeurIPS}}, 2014.

\bibitem[Krishna et~al.(2017)]{krishna2017visual}
Ranjay Krishna et~al.
\newblock Visual genome: Connecting language and vision using crowdsourced
  dense image annotations.
\newblock \emph{IJCV}, 2017.

\bibitem[Kuhn(1955)]{kuhn1955hungarian}
Harold~W Kuhn.
\newblock The hungarian method for the assignment problem.
\newblock \emph{Naval research logistics quarterly}, 1955.

\bibitem[Kusner et~al.(2015)]{kusner2015word}
Matt Kusner et~al.
\newblock From word embeddings to document distances.
\newblock In \emph{ICML}, 2015.

\bibitem[LeCun et~al.(1999)]{lecun1999object}
Yann LeCun et~al.
\newblock Object recognition with gradient-based learning.
\newblock In \emph{Shape, contour and grouping in computer vision}. 1999.

\bibitem[Lee et~al.(2018)]{SCAN}
Kuang-Huei Lee et~al.
\newblock Stacked cross attention for image-text matching.
\newblock In \emph{ECCV}, 2018.

\bibitem[Li et~al.(2019{\natexlab{a}})Li, Duan, Fang, Jiang, and
  Zhou]{li2019unicoder}
Gen Li, Nan Duan, Yuejian Fang, Daxin Jiang, and Ming Zhou.
\newblock Unicoder-{VL}: {A} universal encoder for vision and language by
  cross-modal pre-training.
\newblock \emph{arXiv preprint arXiv:1908.06066}, 2019{\natexlab{a}}.

\bibitem[Li et~al.(2019{\natexlab{b}})Li, Zhang, Li, Li, and Fu]{li2019visual}
Kunpeng Li, Yulun Zhang, Kai Li, Yuanyuan Li, and Yun Fu.
\newblock Visual semantic reasoning for image-text matching.
\newblock In \emph{Proceedings of the IEEE International Conference on Computer
  Vision}, pages 4654--4662, 2019{\natexlab{b}}.

\bibitem[Li et~al.(2020)Li, Yin, Li, Hu, Zhang, Zhang, Wang, Hu, Dong, Wei,
  et~al.]{li2020oscar}
Xiujun Li, Xi~Yin, Chunyuan Li, Xiaowei Hu, Pengchuan Zhang, Lei Zhang, Lijuan
  Wang, Houdong Hu, Li~Dong, Furu Wei, et~al.
\newblock Oscar: Object-semantics aligned pre-training for vision-language
  tasks.
\newblock \emph{ECCV}, 2020.

\bibitem[Lin et~al.(2014)]{lin2014MSCOCO}
Tsung-Yi Lin et~al.
\newblock Microsoft coco: Common objects in context.
\newblock In \emph{ECCV}, 2014.

\bibitem[Liu et~al.(2019)Liu, Mao, Liu, Zhang, Wang, and Zhang]{liu2019focus}
Chunxiao Liu, Zhendong Mao, An-An Liu, Tianzhu Zhang, Bin Wang, and Yongdong
  Zhang.
\newblock Focus your attention: A bidirectional focal attention network for
  image-text matching.
\newblock In \emph{Proceedings of the 27th ACM International Conference on
  Multimedia}, pages 3--11, 2019.

\bibitem[Liu et~al.(2018)]{liu2018scene}
Yishu Liu et~al.
\newblock Scene classification using hierarchical wasserstein cnn.
\newblock \emph{IEEE Transactions on Geoscience and Remote Sensing}, 2018.

\bibitem[Lu et~al.(2019)Lu, Batra, Parikh, and Lee]{lu2019vilbert}
Jiasen Lu, Dhruv Batra, Devi Parikh, and Stefan Lee.
\newblock Vil{BERT}: Pretraining task-agnostic visiolinguistic representations
  for vision-and-language tasks.
\newblock In \emph{NeurIPS}, 2019.

\bibitem[Malinowski and Fritz(2014)]{malinowski2014multi}
Mateusz Malinowski and Mario Fritz.
\newblock A multi-world approach to question answering about real-world scenes
  based on uncertain input.
\newblock In \emph{{NeurIPS}}, pages 1682--1690, 2014.

\bibitem[Nam et~al.(2017)Nam, Ha, and Kim]{nam2017dual}
Hyeonseob Nam, Jung-Woo Ha, and Jeonghee Kim.
\newblock Dual attention networks for multimodal reasoning and matching.
\newblock In \emph{CVPR}, pages 299--307, 2017.

\bibitem[Neubeck and Van~Gool(2006)]{neubeck2006efficient}
Alexander Neubeck and Luc Van~Gool.
\newblock Efficient non-maximum suppression.
\newblock In \emph{18th International Conference on Pattern Recognition
  (ICPR'06)}, volume~3, pages 850--855. IEEE, 2006.

\bibitem[Niu et~al.(2017)]{niu2017hierarchical}
Zhenxing Niu et~al.
\newblock Hierarchical multimodal lstm for dense visual-semantic embedding.
\newblock In \emph{ICCV}, 2017.

\bibitem[Peyr{\'e} and Cuturi(2017)]{peyre2017computational}
Gabriel Peyr{\'e} and Marco Cuturi.
\newblock Computational optimal transport.
\newblock Technical report, 2017.

\bibitem[Plummer et~al.(2015)]{plummer2015flickr30k}
Bryan~A Plummer et~al.
\newblock Flickr30k entities: Collecting region-to-phrase correspondences for
  richer image-to-sentence models.
\newblock In \emph{ICCV}, pages 2641--2649, 2015.

\bibitem[Qiao et~al.(2019)]{qiao2019mirrorgan}
Tingting Qiao et~al.
\newblock Mirrorgan: Learning text-to-image generation by redescription.
\newblock \emph{CVPR}, 2019.

\bibitem[Reed et~al.(2016)]{reed2016generative}
Scott Reed et~al.
\newblock Generative adversarial text to image synthesis.
\newblock \emph{ICML}, 2016.

\bibitem[Ren et~al.(2015)Ren, He, Girshick, and Sun]{ren2015faster}
Shaoqing Ren, Kaiming He, Ross Girshick, and Jian Sun.
\newblock Faster r-cnn: Towards real-time object detection with region proposal
  networks.
\newblock In \emph{{NeurIPS}}, pages 91--99, 2015.

\bibitem[Rubner et~al.(2000)Rubner, Tomasi, and Guibas]{rubner2000earth}
Yossi Rubner, Carlo Tomasi, and Leonidas~J Guibas.
\newblock The earth mover's distance as a metric for image retrieval.
\newblock \emph{International journal of computer vision}, 40\penalty0
  (2):\penalty0 99--121, 2000.

\bibitem[Schuster and Paliwal(1997)]{schuster1997bidirectional}
Mike Schuster and Kuldip~K Paliwal.
\newblock Bidirectional recurrent neural networks.
\newblock \emph{IEEE Transactions on Signal Processing}, 45\penalty0
  (11):\penalty0 2673--2681, 1997.

\bibitem[Su et~al.(2019)Su, Zhu, Cao, Li, Lu, Wei, and Dai]{su2019vl}
Weijie Su, Xizhou Zhu, Yue Cao, Bin Li, Lewei Lu, Furu Wei, and Jifeng Dai.
\newblock {VL-BERT}: {P}re-training of generic visual-linguistic
  representations.
\newblock \emph{arXiv preprint arXiv:1908.08530}, 2019.

\bibitem[Sun et~al.(2019)Sun, Myers, Vondrick, Murphy, and
  Schmid]{sun2019videobert}
Chen Sun, Austin Myers, Carl Vondrick, Kevin Murphy, and Cordelia Schmid.
\newblock Video{BERT}: {A} joint model for video and language representation
  learning.
\newblock \emph{ICCV}, 2019.

\bibitem[Tan and Bansal(2019)]{tan2019lxmert}
Hao Tan and Mohit Bansal.
\newblock {LXMERT}: {L}earning cross-modality encoder representations from
  transformers.
\newblock \emph{EMNLP}, 2019.

\bibitem[Vendrov et~al.(2016)Vendrov, Kiros, Fidler, and
  Urtasun]{vendrov2015order}
Ivan Vendrov, Ryan Kiros, Sanja Fidler, and Raquel Urtasun.
\newblock Order-embeddings of images and language.
\newblock In \emph{ICLR}, 2016.

\bibitem[Villani(2008)]{villani2008optimal}
C{\'e}dric Villani.
\newblock \emph{Optimal transport: old and new}, volume 338.
\newblock Springer Science \& Business Media, 2008.

\bibitem[Vinyals et~al.(2015)]{vinyals2015show}
Oriol Vinyals et~al.
\newblock Show and tell: A neural image caption generator.
\newblock In \emph{CVPR}, pages 3156--3164, 2015.

\bibitem[Wang et~al.(2016)]{wang2016learning}
Liwei Wang et~al.
\newblock Learning deep structure-preserving image-text embeddings.
\newblock In \emph{CVPR}, 2016.

\bibitem[Wang et~al.(2019)Wang, Yang, Qian, Ma, Lu, Li, and
  Fan]{wang2019position}
Yaxiong Wang, Hao Yang, Xueming Qian, Lin Ma, Jing Lu, Biao Li, and Xin Fan.
\newblock Position focused attention network for image-text matching.
\newblock \emph{arXiv preprint arXiv:1907.09748}, 2019.

\bibitem[Xie et~al.(2018)Xie, Wang, Wang, and Zha]{xie2018fast}
Yujia Xie, Xiangfeng Wang, Ruijia Wang, and Hongyuan Zha.
\newblock A fast proximal point method for computing exact wasserstein
  distance.
\newblock \emph{arXiv preprint arXiv:1802.04307}, 2018.

\bibitem[Xu et~al.(2015)]{xu2015show}
Kelvin Xu et~al.
\newblock Show, attend and tell: Neural image caption generation with visual
  attention.
\newblock In \emph{ICML}, pages 2048--2057, 2015.

\bibitem[Yu et~al.(2017)]{yu2017multi}
Dongfei Yu et~al.
\newblock Multi-level attention networks for visual question answering.
\newblock In \emph{CVPR}, 2017.

\bibitem[Zhao et~al.(2018)Zhao, Li, Zhao, and Feng]{zhao2018weakly}
Fang Zhao, Jianshu Li, Jian Zhao, and Jiashi Feng.
\newblock Weakly supervised phrase localization with multi-scale anchored
  transformer network.
\newblock In \emph{Proceedings of the IEEE Conference on Computer Vision and
  Pattern Recognition}, pages 5696--5705, 2018.

\bibitem[Zheng et~al.(2017)]{zheng2017dual}
Zhedong Zheng et~al.
\newblock Dual-path convolutional image-text embedding with instance loss.
\newblock \emph{arXiv}, 2017.

\end{thebibliography}
}
\clearpage

\end{document}